\documentclass{article}

    \usepackage[nonatbib,preprint]{neurips_2019}

\usepackage[utf8]{inputenc} 
\usepackage[T1]{fontenc}    
\usepackage{hyperref}       
\usepackage{url}            
\usepackage{booktabs}       
\usepackage{makecell}
\usepackage{multirow}
\usepackage{amsfonts, amsmath}       
\usepackage{amssymb}
\usepackage{bbm}
\usepackage{nicefrac}       
\usepackage{microtype}      

\newcommand{\RNum}[1]{\uppercase\expandafter{\romannumeral #1\relax}}

\usepackage{enumitem}

\usepackage{pgfplots}
\usepackage{tikz}
\tikzset{ font=\textbf{}{\fontsize{9pt}{12}\selectfont}}
\usepackage{subcaption}
\usepackage{titlesec}
\usepackage{flexisym}
\title{Blocksworld Revisited: Learning and Reasoning to Generate Event-Sequences from Image Pairs}

\author{%
  Tejas Gokhale,\quad Shailaja Sampat,\quad Zhiyuan Fang,\quad Yezhou Yang,\quad Chitta Baral\\
   Arizona State University \\
   \texttt{\{tgokhale, ssampa17, zfang29, yz.yang, chitta\}@asu.edu}
}

\begin{document}
\maketitle
\begin{abstract}
The process of identifying changes or transformations in a scene along with the ability of reasoning about their causes and effects, is a key aspect of intelligence.
In this work we go beyond recent advances in computational perception, and introduce a more challenging task, Image-based Event-Sequencing (IES).
In IES, the task is to predict a sequence of actions required to rearrange objects from the configuration in an input source image to the one in the target image.
IES also requires systems 
to possess inductive generalizability.
Motivated from evidence in cognitive development, we compile the first IES dataset, the Blocksworld Image Reasoning Dataset (BIRD)
\footnote{BIRD is available publicly at \href{https://asu-active-perception-group.github.io/bird_dataset_web/}{https://asu-active-perception-group.github.io/bird\_dataset\_web/}}
which contains images of wooden blocks in different configurations, and the sequence of moves to rearrange one configuration to the other.
We first explore the use of existing deep learning architectures and show that these end-to-end methods under-perform in inferring temporal event-sequences and fail at inductive generalization.
We then propose a modular two-step approach: Visual Perception followed by Event-Sequencing, and demonstrate improved performance by combining learning and reasoning.
Finally, by showing an extension of our approach on natural images, we seek to pave the way for future research on event sequencing for real world scenes.
\end{abstract}

\section{Introduction}
\label{intro}
Deep neural networks trained in an end-to-end fashion have resulted in exceptional advances in computational perception, especially in object detection \cite{he2017mask,redmon2016you}, semantic segmentation \cite{chen2018encoder,zhao2017pyramid}, and action recognition \cite{carreira2017quo}.
Given this capability, a next step is to enable vision modules to reason about perceived visual entities such as objects and actions.
Some works {\cite{zhou2018temporal}}
approach this paradigm by inferring spatial, temporal and semantic relationships between the entities.
Other works deal with identifying changes in these relationships (spatial {\cite{jhamtani2018learning}} or temporal {\cite{park2019viewpoint}}).
Spatial reasoning has been explored in the context of Visual Question Answering (VQA) via the CLEVR dataset \cite{clevr}.
Relation Networks (RN) proposed in {\cite{relational}} augment image feature extractors and language embedding modules with a composite and differentiable relational reasoning module, to answer questions about attributes and relative locations of blocks.

In this work, we go beyond and present a new task, Image-based Event Sequencing (IES).
Given a pair of images, the goal in IES is to predict a temporal sequence of events or moves needed to rearrange the object-configuration in the first image to that in the second.
An important requirement for potential IES solvers is {\it inductive generalizability}, the ability of predicting an event-sequence of any length, even when trained only on samples with shorter lengths.
A simple analogy can be found in the process of sorting a list; a correct program should be able to sort irrespective of the number of swaps required. Inductive generalizability is a characteristic possessed by humans; a person who knows how to drive, but has never driven more than 20 miles, can drive to any farther destination reachable by road, provided with the correct directions.

To validate IES systems, we need a testbed, and to the best of our knowledge, no public testbed exists (with detailed annotations about spatial configurations and event-sequences).
While CLEVR {\cite{clevr}} and Sort-of-CLEVR {\cite{relational}} also contain images of block-configurations, they are artificially generated and more importantly do not include detailed sequences between pairs of images.
Moreover, the blocks in these datasets are never stacked or in contact and so there are no constraints on movement of these blocks.
The creators of these datasets force the blocks to be at a minimum margin from each other, and thus any block can be picked up and moved without affecting the other blocks in the configuration. 
However in real world scenes, objects do impose constraints on one another, for instance a book which has a cup on top of it, cannot be moved without disturbing the cup.
Thus, we compile the Blocksworld Reasoning Image Dataset (BIRD)
that includes 1 million samples containing a source image and a target image (each containing wooden blocks arranged in different configurations), and all possible sequences of moves to rearrange the source configuration into the target configuration.

To tackle the IES challenge, we propose a modular approach and decompose the problem into two stages, Visual Perception and Event-Sequencing.
Stage-{\RNum{1}} is an encoder network that converts each input image into a vector representing the spatial and object-level configuration of the image.
Stage-{\RNum{2}} uses these vectors to generate event-sequences. 
This decomposition of the system into two modules makes the sequencing module standalone and reproducible.
While the encoder can change based on domain, the sequencing module once learned on the blocksworld domain, can be reused on more complex domains, such as real-world scenes.
We compare this two-stage approach with several existing end-to-end baselines, and show significant improvement.

To test for inductive generalization, we train our models on data containing true sequences with an upper bound on length, and test them on samples that require sequences of longer lengths.
We observe that end-to-end methods fail to generalize while two-stage methods exhibit inductive capabilities.
Inductive Logic Programming {\cite{muggleton1991inductive}} which combines learning and reasoning by using background knowledge, performs the best under this setting, and can be used to learn event-sequences with unbounded lengths.

Thus, our contributions are fourfold; we:
\begin{enumerate}[noitemsep,topsep=0pt]
\item introduce the first IES challenge and compile the BIRD dataset as a testbed, 
\item show that end-to-end training fails at event-sequence generation and inductive generalization,
\item show the benefits of a two-stage approach, and
\item show that a sequencing module learned on the BIRD data can be re-used on natural images, yielding a capability towards human level intelligence {\cite{sarama:bblocks}}.
\end{enumerate}

\section{Related Work}
\label{related}
We identify three tasks that are most relavant to the IES task; ``Spot-the-Difference", Reasoning in Visual Question Answering (VQA) and Visual Relationship Extraction.

\subsection{Spot-the-Difference}
Change detection between a pair of images has been explored previously with image differencing techniques using unsupervised {\cite{bruzzone2000automatic}} or semi-supervised {\cite{gueguen2015large}} methods. However, these are pixel-level techniques and are not designed to compute the semantic differences between two images.  The ``Spot-the-Difference" task introduced in {\cite{jhamtani2018learning}} leverages natural language annotations to generate multi-sentence description of differences between two images. An existing work which comes closest to our IES task is the Viewpoint Invariant Change Captioning (VICC) task {\cite{park2019viewpoint}} where the aim is to generate a textual description of the changes between objects in two images ({\it before} and {\it after}). However, the VICC model only predicts {\it which} object in the before-image changed position, but does not specify its position in the after-image or how that change might have taken place. Since the VICC model is built on the CLEVR dataset - in which blocks are never in contact with each other, there are no constraints on movement, therefore making the reasoning aspect of VICC simpler than IES.

\subsection{Reasoning in Visual Question Answering}
Spatial reasoning has been explored extensively in the context of Visual Question Answering (VQA) via the CLEVR dataset.  Given an image, the task is to answer questions that require reasoning about attributes such as shapes, textures, colors and relative locations of objects in the image. Relation Networks (RN) proposed in {\cite{relational}} seek to solve this problem by augmenting image feature extractors and language embedding modules with a relational reasoning module. The RN is an end-to-end differentiable and composite function that computes relations between the question embedding and all possible combinations of image features. In comparison, the IES task requires not only understanding attributes of objects, but also requires inferring a sequence of actions or events that could lead to a desired configuration of the objects. In VQA, the input is an image-question pair and the output is a single word or class label, whereas in IES, the input is an image-image pair and the output is a temporal sequence of events. The IES task can be thought of as the question - ``How would you navigate from the source image to the target image in blocksworld?". Although the outcomes in VQA and IES are of different types, the capabilities required to perform inference involve building a certain level of reasoning to perform spatial and relational tasks.

\subsection{Visual Relationship Extraction}
Another category related to the IES task is visual relationship extraction in which the aim is to embed objects and their relationships into {\it \texttt{<}subject, relation, object\texttt{>}} triplets, given an image and its caption.
{\cite{plummer2017phrase}} consider each triplet as a separate class and learn to predict each relationship as a triplet as well as localize it as a bounding box in image-space, while {\cite{elhoseiny}} allow a continuous output space for objects and relations. 

\section{Image-based Event Sequencing (IES)}
\label{problem}
In this section, we formulate the IES task in terms of inputs, outputs and desired properties of the systems that attempt the task.

\begin{figure}[!t]
    \centering
    \includegraphics[width=\linewidth]{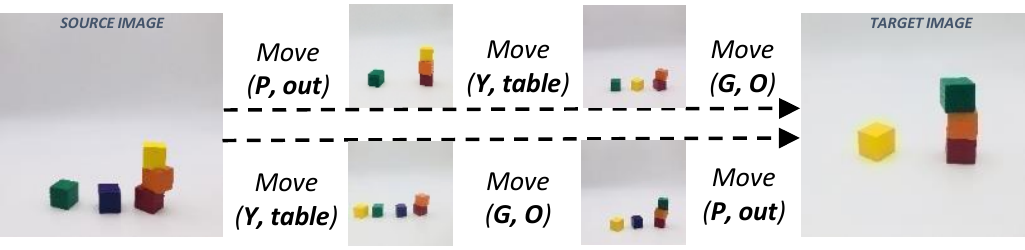}
    \caption{Illustration of two event-sequences between an image-pair (with intermediate configurations shown for clarity).}
    \label{fig:sequencing}
\end{figure}

\subsection{Problem Statement}
The input to the IES task is a pair of images ({\it source}  $\mathnormal{I^S}$ and {\it target} $\mathnormal{I^T}$), that contain objects appearing in different configurations.
The goal of the IES task is to find an event sequence $\mathnormal{M} = [\mathnormal{m_1}, \dots, \mathnormal{m_L}]$, such that performing $\mathnormal{M}$ on $\mathnormal{I^S}$ leads to the configuration in $\mathnormal{I^T}$.
Here L is the length of sequence M
and $\mathnormal{m_t}$ is the move at time $\mathnormal{t} \in \{1,\dots,L\}$.
Figure \ref{fig:sequencing} shows an example. Note that a pair of images can have multiple, unique or no permissible event-sequences. 

\subsection{Inductive Generalization}
Under this problem setting, we define the concept of inductive generalization. Given a training dataset $\mathcal{S}$ with $\mathnormal{n}$ samples, let $L_{max}$ be the maximum length of sequences found in the dataset.
\begin{align}
    \mathcal{S} &= \{X_1 \dots X_n\} \qquad \text{where} \quad X_i = (I_i^S, I_i^T, M_i)\quad \forall i \in \{1,\dots,n\}\\
    L_{max}     &= \max_{i \in \{1, \dots, n\}} |M^i|
\end{align}
Then, a system is said to possess inductive generalizability if it is able to predict event-sequences accurately for any sample $\hat{X} = (\hat{I}^S, \hat{I}^T, \hat{M})$ where $|\hat{M}| > L_{max}$.

\section{Blocksworld Image Reasoning Dataset (BIRD)}
\label{dataset}
In this work, we focus on the ``Blocksworld" setting where every image contains blocks of different colors arranged in various configurations.

\subsection{Motivation for BIRD}
What's so special about blocks? Our motivation for constructing a curated dataset of blocksworld images comes from literature in cognitive development.
Extensive studies such as {\cite{piaget:play,harriet:art, sally:bblocks}} show that playing with wooden blocks benefits the early stages of a child's development.
These works show how block-play aids in development of a child's sensorimotor, symbolic, logical, mathematical as well as abstract and causal reasoning abilities.
{\cite{sarama:bblocks}} have argued that building with blocks enables children to {\it mathematize} the world around them in terms of physics, geometry, visual attributes, and abstract semantics or meanings assigned to blocks.

The crucial insight from these works is that the task of reasoning about a complex visual scene benefits from abstractions in terms of blocks; when every object in a scene is treated as a block, the entire scene can be re-imagined in the blocksworld framework.
Correctly generating event-sequences from images requires perceiving objects, colors, textures and reasoning about spatial relationships in order to come up with a plan to build towards the goal.
{\cite{gupta2010blocks} use an ``Interpretation-by-Synthesis" approach to progressively build up representations of images.
We propose a similar construct for visual perception that could aid in reasoning tasks such as the one in IES.

With the claim that the IES task can be learned on the Blocksworld domain, and extended and reused on other domains without re-training, we introduce a new dataset -- the {\it Blocksworld Image Reasoning Dataset (BIRD)}.

\subsection{Constructing BIRD}
\subsubsection{Image Capturing}
\pgfplotstableread[row sep=\\,col sep=&]{
    interval & towers & blocks \\
    0   &   1       &   1 \\
    1   &   6       &   1236 \\
    2   &   60      &   2310 \\
    3   &   360     &   1920 \\
    4   &   1440    &   1080 \\
    5   &   3600    &   720 \\
    }\mydata
    
\begin{figure}[t]
\centering 
  \begin{subfigure}{0.49\linewidth}
  \begin{tikzpicture}
    \begin{axis}[ybar,
            enlarge y limits={upper, value=0.5},
            symbolic x coords={0,1,2,3,4,5},
            nodes near coords,
            width=\linewidth,height=3cm,
            ylabel= Images,
            ylabel style={at={(4ex,0.5)}},
            yticklabels={\empty},
            xtick style={draw=none},
            ytick style={draw=none}
            ]
        \addplot table[x=interval,y=towers]{\mydata};
    \end{axis}
\end{tikzpicture}
    \caption{Based on number of stacks} \label{fig:M1}  
  \end{subfigure}
  \begin{subfigure}{0.49\linewidth}
  \begin{tikzpicture}
    \begin{axis}[ybar,
            enlarge y limits={upper, value=0.5},
            symbolic x coords={0,1,2,3,4,5},
            nodes near coords,
            width=\linewidth,height=3cm,
            ylabel= Images,
            ylabel style={at={(4ex,0.5)}},
            yticklabels={\empty},
            xtick style={draw=none},
            ytick style={draw=none}
            ]
        \addplot table[x=interval,y=blocks]{\mydata};
    \end{axis}
\end{tikzpicture}
    \caption{Based on number of blocks in image} \label{fig:M2}  
  \end{subfigure}
\caption{Distribution of block images in BIRD}
\label{fig:img_dataset}
\end{figure}
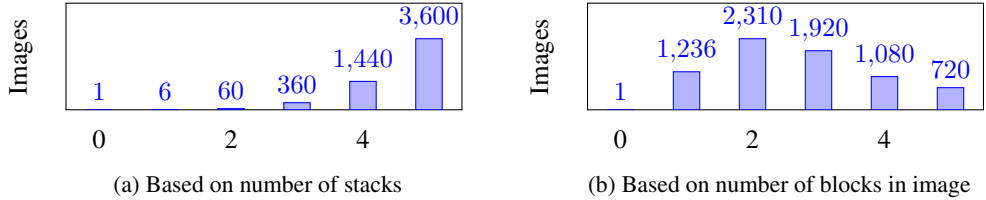

BIRD consists of 7267 images of blocks arranged in different configurations that we captured in white background and uniform lighting conditions. 
We use wooden blocks from a set of six colors $\mathcal{C}$  and arrange them in all possible permutations.
In doing so, we follow two constraints -- an image contains no more than five blocks, and no two blocks of the same color.
Figure \ref{fig:img_dataset} shows the distribution of the dataset based on number of blocks and stacks of blocks in the image.
Our intention is to distinguish our dataset from CLEVR \cite{clevr} in that our dataset contains blocks that are in contact or stacked on top of each other, and also that we use real images (as opposed to rendered images in CLEVR).

\subsubsection{Annotation}
\begin{figure}[t]
    \centering
 \includegraphics[width=0.66\linewidth]{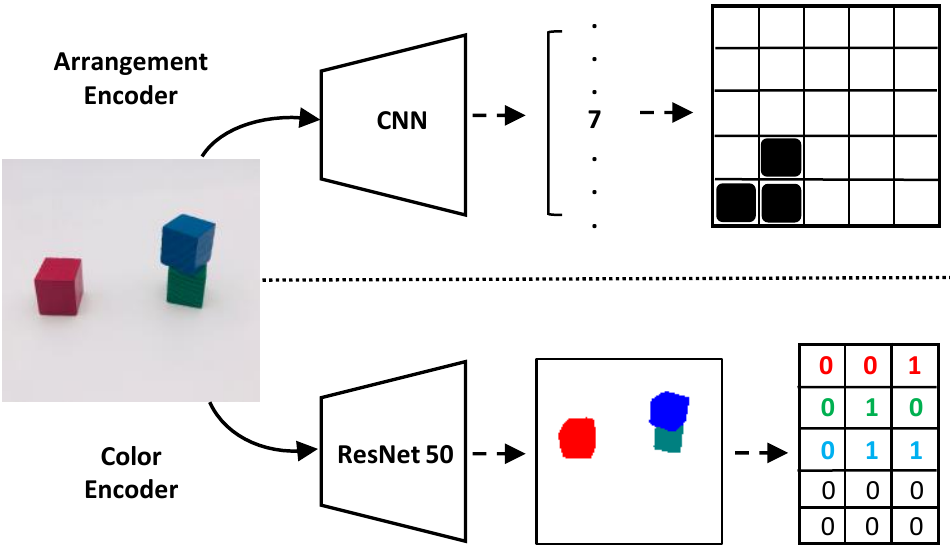}
 \caption{Images with their arrangement and color vector}
 \label{fig:encoding}
\end{figure}
We annotate each image with two vectors that uniquely represent the configuration of blocks as shown in Figure {\ref{fig:encoding}}.
The ``{\it color-blind} arrangement vector" represents the locations of blocks in a grid. 
The ``color vector" represents colors of the blocks from bottom-to-top and left-to-right, with each color represented as a 3-bit binary vector.
For every pair of source and target images, we assign all possible minimal-length event-sequences $\mathnormal{M}$, with each move $\mathnormal{m_t}$ in the sequence given by:
\begin{equation} \label{eq:moves}
    \begin{split}
    \qquad &move(\mathbf{X}, \mathbf{Y}, \mathnormal{t}) ;  t \in \{0, 1, \dots, 7\}, \mathbf{X} \neq \mathbf{Y}\\
        \text{where } &\mathbf{X}\in \mathcal{C} , \mathbf{Y}\in \mathcal{D} = \mathcal{C} \cup \{``table"\} \cup \{``out"\}.
    \end{split}
\end{equation}

For example, $move(\mathbf{R}, \mathbf{G}, \mathnormal{2})$ implies that a red block is moved on top of a green block at the second time-step.
We pair every image in the dataset with every other image and use the {\sc clingo} {\cite{gebser2011potassco}} Answer Set Programming solver to generate a dataset of {\it $\langle$image-image-sequence$\rangle$} triplets as shown in Figure {\ref{fig:sequencing}}, uniformly sampled across all sequence lengths $\ell \in \{\text{no-sequence}, 0, \dots, 8\}$.
The maximum length of minimal-length sequences in our dataset is 8.

\subsubsection{Background Knowledge}
To reason about the configurations, we use the following background knowledge to delineate the conditions under which each move is legal:

\noindent\fbox{%
\parbox{\linewidth}{%
\textbf{Exogeneity}: Block $\mathbf{A}$ can be moved at time t $\iff$ it exists in the configuration $\forall$ $\hat{t}<t$.\\
\textbf{Freedom of Blocks}:
    \begin{enumerate}[noitemsep,topsep=0pt]
        \item Block $\mathbf{A}$ is {\it free} at time t $\Leftrightarrow \forall \mathbf{B}, \neg on(\mathbf{B}, \mathbf{A}, t)$.
        \item Block $\mathbf{A}$ can be moved $\Leftrightarrow$ A is free.
        \item Block $\mathbf{B}$ can be placed on block $\mathbf{A}$ $\Leftrightarrow$ $\mathbf{A}$ is free.
        \item A block that is ``out of table" cannot be moved.
    \end{enumerate}
\textbf{Inertia}: A block unless moved doesn't change location.\\
\textbf{Sequentialism}: At most one move can be performed at each time instance.
}}

\section{Methods}
\label{method}
Armed with our novel dataset, we test two approaches to attempt the Image-based Event Sequencing (IES) task, End-to-End Learning and Modular Two-Stage Methods.

\subsection{End-to-End Learning}
In BIRD, each move is represented according to Equation \ref{eq:moves}. 
Since $|\mathcal{D}| = 8$, we represent each X or Y with a 8-bit one-hot vector, and therefore get a 16-bit representation for each move $m_t$. 
The maximum number of moves for any image-pair in our dataset is 8; therefore our ground-truth event sequence is a 128-bit binary vector.
Our input is a pair of RGB images ($I^S, I^T$); i.e. a 6 channel input with dimensions $256\times256$.
Thus our end-to-end modules are given by:
\begin{equation}
    f_E : \mathbb{R}^{256\times256\times6} \rightarrow \mathnormal{\{0, 1\}}^{128} .
    \label{eq:e2e}
\end{equation}

We train deep neural network architectures that can leverage spatial context such as Resnet-50 {\cite{he2016deep}}, PSPNet {\cite{zhao2017pyramid}} and Relational Networks (RN) {\cite{relational}}, to directly generate event-sequences from image pairs.
We use Pyramid Scene Parsing network (PSPNet) as a baseline since it uses pyramidal pooling as a global contextual prior for extracting spatial relations.
It is worth exploring if spatial relationships captured by PSPNet for semantic segmentation can be useful in the IES task. 
Relational Networks have been shown to work for relational reasoning in Visual Question Answering and have an image-question pair as input. 
An RN extracts image features using a Convolutional Neural Network (CNN) \cite{lecun1998gradient} and text features using Long-Short Term Memory (LSTM) \cite{hochreiter1997long} embedding and uses these features as inputs to the Relational Module. 
In our case, we have an image-image pair instead thus, we replace the LSTM with another CNN feature extractor and train the RN end-to-end.

\subsection{Modular Methods}
We decompose the task into Stage-\RNum{1} (Visual Perception) and Stage-\RNum{2} (Event Sequencing).

\subsubsection{Stage-\RNum{1}: Visual Perception}
Stage-\RNum{1} is trained to encode input images into an interpretable representation.
We identify that spatial localization of blocks with respect to one another requires knowing {\it where} the relative location of each block, given by an {\it arrangement vector}, along with the {\it characteristics} of each block, given by a {\it color vector}.
We train a 8-layer convolutional network ($f_A$) to encode this arrangement vector.
In our dataset, the maximum number of blocks is 5, so the arrangement can be expressed as a $5\times5$ grid. 
Then we train a Resnet-50 based color grounding module ($f_C$) as in \cite{fang2019modularized}, and use it along with the predicted arrangement vector to obtain the color vector, that represents the color of each block as a 3-bit binary vector, in a bottom-to-top, left-to-right order.
Thus our visual perception is given by the two encoders, expressed as:
\begin{equation}
    \mathnormal{f_A} : \mathbb{R}^{256\times256\times3} \rightarrow \mathbb{R}^{5\times5} , \mathnormal{f_C} : \mathbb{R}^{256\times256\times3} \rightarrow \mathbb{R}^{5\times3} .\label{eq:enc}
\end{equation}

\subsubsection{Stage-\RNum{2}: Event Sequencing}
Stage-\RNum{2} is trained to use the encoded representation of images to generate minimal-length sequences of moves to reach the target from the source configuration.
We compare the efficacy of Fully Connected Neural Networks (FC), reinforcement learning using the Q-Learning algorithm (QL) and rule-based Inductive Logic Programming (ILP). 
The worst case sequence length (8) serves as upper bound for sequence generation using QL and ILP. 
Given an action $m_t$ and a configuration $z_t$, we also develop a {\it Logic Engine} that can deterministically generate the next configuration $z_{t+1}$. 
The logic engine ($g_{l}$) can be expressed as:
    \begin{align}
        \mathnormal{z_{t+1}} &= \mathnormal{g_{l}(z_{t}, m_t)},  \\
        \text{where} \quad \mathnormal{z_0} &=  \lbrack f_A(I^S), f_C(I^S) \rbrack .
    \end{align}

\textbf{Deep Neural Networks.}
We explore if conventional neural networks can be used to generate discrete event-sequences in the IES task, by using Fully Connected (FC) networks as one of the baselines to predict event-sequences.
We train an FC network with five layers under a multi-label classification paradigm with binary cross-entropy loss and Adam optimizer.

\textbf{Reinforcement Learning.}
Q-learning (QL) {\cite{watkins1992q}} is a widely used model-free reinforcement learning algorithm that models the world as a finite Markov Decision Process (MDP) in which agents receive a reward based on the action they perform at every time-step.
The QL algorithm finds an optimal policy by total discounted expected reward.
We look at our event-sequencing problem as a finite MDP between the start image and target image with moves in the temporal sequence being analogous to ``actions".
The policy that we learn in this reinforcement learning framework is designed such that it is consistent with the background knowledge for event-sequencing in BIRD.

\textbf{Inductive Logic Programming.}
Inductive Logic Programming (ILP) {\cite{muggleton1991inductive}} is a subclass of machine learning algorithms that aims to learn logic programs.
Given the encoded background knowledge $\mathcal{B}$ of the domain  and a set of positive and negative examples represented as structured facts $\mathcal{E}^{+}$ and $\mathcal{E^{-}}$, the ILP system learns a logic program that entails $\mathcal{E^{+}}$ but not $\mathcal{E^{-}}$.

{\cite{mitra2016addressing}} have shown that the addition of a formal reasoning layer to standard statistical machine learning approaches significantly increases the reasoning capability of an agent.
With that motivation, we use ILP to learn Answer Set Programs for our event-sequencing task.
Using examples from BIRD represented in a structured ASP format, we learn effects of the action $move(X, Y, t)$ on the relative positions of X, Y equivalent to the rule:
\begin{equation}
    on(X,Y,t+1) \ {:}\textrm{-} \ move(X,Y,t).
\end{equation}

\section{Experiments}
\label{experiments}
\subsection{Evaluation Metrics}
We define two metrics for our experiments.
If $\mathnormal{y}$ and $\hat{y}$ are the  and the predicted sequence, then {\it Full Sequence Accuracy} (FSA) is the percentage of exact matches, and {\it Step Level Accuracy }(SLA) is the percentage of common moves between $\mathnormal{y}$ and $\hat{y}$.

\begin{align}
    \text{FSA} &= \frac{1}{N} \sum_{i=1}^{N}\mathbbm{1}\{y^i == \hat{y}^i \} \\
    \text{SLA} &= \frac{1}{N}\sum_{i=1}^{N}\frac   {\sum_{l=1}^{L} {\mathbbm{1}\{y_\ell^i == \hat{y_\ell}^i \} }}{L} \label{eq:accuracy_precision}
\end{align}

\subsection{Results on BIRD}
\begin{table}[t]
  \caption{Comparison of all methods for the IES task on BIRD, with respect to the FSA and SLA metrics. Note that {\it PR} refers to Stage-\RNum{1} with perfect recognition.}
  \label{tab:blocksworld_baseline}
  \centering
  \begin{tabular}{llcc}
    \toprule
    \textbf{Approach} & \textbf{Method} & \textbf{FSA} & \textbf{SLA} \\
    \midrule
    \textbf{Human} & & 100 & 100\\
    \midrule
    \multirow{3}{*}{\textbf{End-to-End Learning}}
        & ResNet50  & 30.52 & 36.26\\
        & PSPNet    & 35.04 & 56.69\\
        & RN        & 34.37 & 52.09\\
    \midrule
    \multirow{3}{*}{\textbf{PR + Stage-\RNum{2}}}
        & FC        & 68.87 & 72.58\\
        & QL        & 84.10 & 87.83\\
        & ILP       & \textbf{100}   & \textbf{100}\\
    \midrule
    \multirow{3}{*}{\textbf{Stage-\RNum{1} + Stage-\RNum{2}}}
        & FC        & 56.25 & 60.24\\
        & QL        & 68.98 & 71.17\\
        & ILP       & \textbf{83.60}   & \textbf{88.53}\\
    \bottomrule
  \end{tabular}
\end{table}

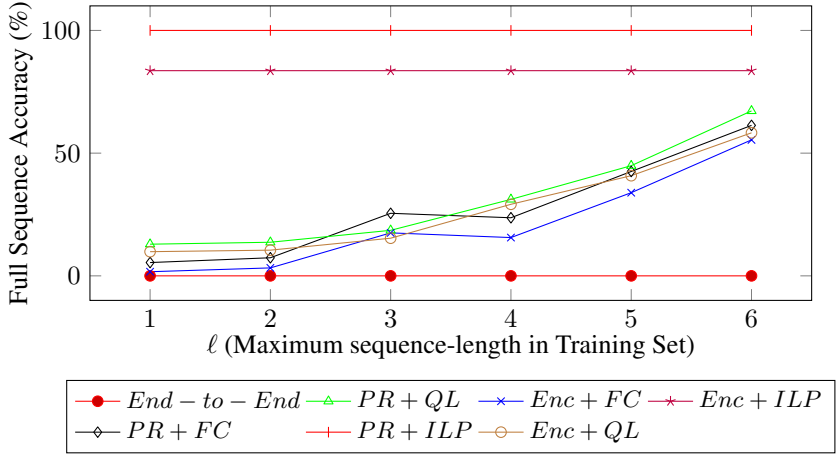
\begin{figure}[t]
\centering
\begin{tikzpicture}
\begin{axis}[
height=5.5cm, width=0.8\linewidth,
xlabel=$\ell$ (Maximum sequence-length in Training Set),
ylabel=Full Sequence Accuracy (\%),
xlabel style={at={(0.5,0.05)}},
ylabel style={at={(2ex,0.5)}},
transpose legend,
legend columns=2,
legend style={at={(0.5, -0.27)},anchor=north},
legend style={font=\small},
legend cell align={left},
cycle multi list={color list\nextlist[1 of]mark list}
]
\addplot coordinates {
(1, 0)
(2, 0)
(3, 0)
(4, 0)
(5, 0)
(6, 0)
}; \addlegendentry{$End-to-End$}
\addplot+[black,line join=round,mark=diamond] coordinates {
(1, 5.42)
(2, 7.37)
(3, 25.49)
(4, 23.67)
(5, 42.53)
(6, 61.31)
}; \addlegendentry{$PR+FC$}
\addplot+[green,line join=round,mark=triangle] coordinates {
(1, 12.90)
(2, 13.69)
(3, 18.55)
(4, 31.16)
(5, 44.89)
(6, 67.21)
}; \addlegendentry{$PR+QL$}
\addplot+[red,line join=round,mark=|] coordinates {
(1, 100)
(2, 100)
(3, 100)
(4, 100)
(5, 100)
(6, 100)
}; \addlegendentry{$PR+ILP$}
\addplot+[blue,line join=round,mark=x] coordinates {
(1, 1.68)
(2, 3.24)
(3, 17.51)
(4, 15.6)
(5, 33.89)
(6, 55.4)
}; \addlegendentry{$Enc+FC$}
\addplot+[brown,line join=round,mark=o] coordinates {
(1, 9.85)
(2, 10.47)
(3, 15.34)
(4, 29.16)
(5, 40.81)
(6, 58.28)
}; \addlegendentry{$Enc+QL$}
\addplot+[purple,line join=round,mark=star] coordinates {
(1, 83.6)
(2, 83.6)
(3, 83.6)
(4, 83.6)
(5, 83.6)
(6, 83.6)
}; \addlegendentry{$Enc+ILP$}
\end{axis}
\end{tikzpicture}
\caption{Inductive capability of each method, shown in terms of FSA on the test set containing sequences longer than those used for training. (Best when viewed in color).}
\label{fig:induction}
\end{figure}

We evaluate and compare end-to-end and two-stage methods in Table \ref{tab:blocksworld_baseline}. 
Two-stage methods significantly outperform all end-to-end methods, even with imperfect Stage-\RNum{1} encoders (Enc).
Since the output space is exponentially large, we postulate that end-to-end networks lack the ability to map from pixel-space to this large sequence-space.

If an image-pair requires more number of moves than present in the training data, our system should inductively infer this longer sequence of steps.
We test this {\it Inductive Generalizability} with an ablation study; we create datasets such that the training set has samples with maximum length $\ell$ and the test set with minimum length $\ell+1$.
Figure \ref{fig:induction} illustrates that end-to-end methods do not possess this ability, while two stage methods generalize well to some degree; as $\ell$ increases, the inductive capability of QL and FC increases. Inductive Logic Programming with perfect recognition (PR) is able to generalize irrespective of the value of $\ell$.

\subsection{Results on Natural Images}
\begin{figure}
    \centering
    \includegraphics[width=\textwidth]{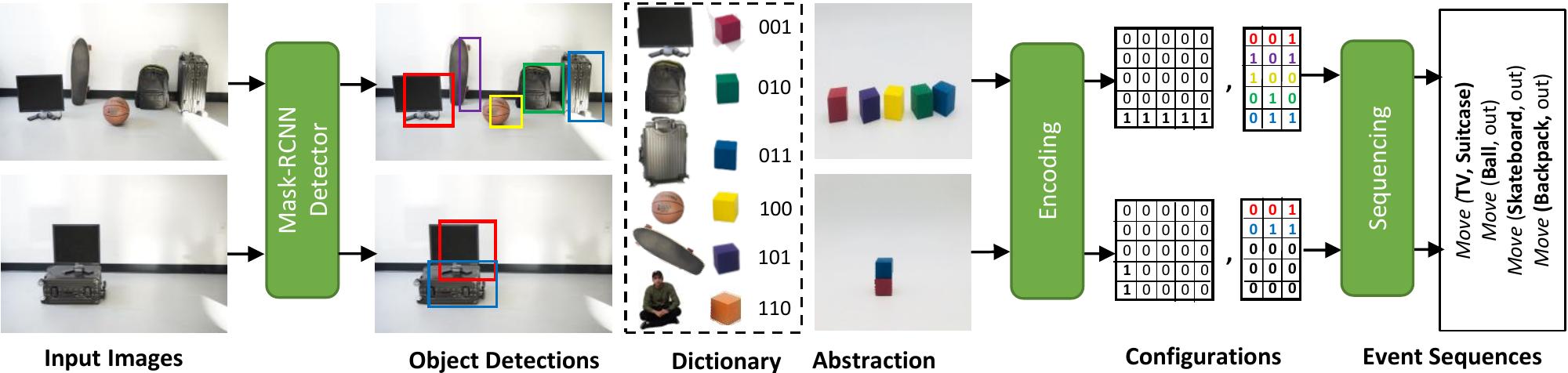}
    \caption{Experiments on Natural Images: Given a source and target image  we get object detections using a Mask-RCNN. These detections are {\it re-imagined} in the blocksworld framework on which we perform event-sequencing using models trained on BIRD to get output moves.}
    \label{fig:natural}
\end{figure}

\begin{table}[t]
    \centering
    \small
    \caption{Results of using BIRD sequencing module for natural images (with Perfect Recognition or Mask-RCNN as Stage-I)}
    \begin{tabular}{lccccccccccc}
    \toprule
        \textbf{Approach} &  \multicolumn{3}{l}{\makecell{\textbf{PR \texttt{+} Stage-\RNum{2}}}} & \multicolumn{3}{l}{\makecell{\textbf{Stage-\RNum{1} \texttt{+} Stage-\RNum{2}}}} \\
        
        & FC & QL & ILP & FC & QL & ILP \\
        \midrule
        FSA (\%) & 55.34 & 92.20 & 100 & 47.47 & 64.26 & 75.55 \\
        SLA (\%) & 61.06 & 96.42 & 100 & 51.71 & 69.16 & 80.57 \\
        \bottomrule
    \end{tabular}
    
    \label{tab:natural_results}
\end{table}

We collected a set of 30 images which contain the object classes ``Person", ``TV", ``Suitcase", ``Table", ``Backpack" and ``Ball" as a prototype to test the hypothesis that the sequencing module trained on BIRD can be reused for natural image inputs.
We used a pre-trained Mask-RCNN {\cite{he2017mask}} network to produce object detections and {\it re-imagined} the image in the blocksworld setting, by using a one-to-one mapping from each object to a block-type in BIRD.
Thus for a pair of natural images, we can test various sequencing modules trained on BIRD by directly using the corresponding blocksworld re-imaginations to generate event-sequences as shown in Figure {\ref{fig:natural}}. Table \ref{tab:natural_results} shows a comparison of our Stage-\RNum{2} baselines.

\subsection{Discussion}
\label{discussion}
Table \ref{tab:blocksworld_baseline} shows that all three end-to-end methods are significantly outperformed by two-stage methods, even when using imperfect encoders from Stage-\RNum{1}.
Our output space consists of 8 moves with each move having 48 possibilities, making the number of possible outputs $48^8 \approx 2.8 \times 10^{13}$.
We postulate that end-to-end networks are incapable at handling an output space as large as in IES, and as a result fail to identify the semantic correspondence between the pixel-space and sequence-space.
Since the two-stage approach is guided by the perception module to encode a interpretable latent vector, it aids the sequencing module to infer sequences.
We argue that encoding scenes from pixel-domain into semantic and interpretable representations and then using these for reasoning has an edge over learning to reason directly from pixels.
ILP with background knowledge outperforms all the other baselines as shown in (Table \ref{tab:blocksworld_baseline}).
We note that while Q-Learning also achieves good accuracies on the IES task, it is not able to generalize as well as ILP in terms of inductive reasoning capabilities as can be seen from Figure \ref{fig:induction}.

\section{Conclusion}
\label{conclusion}
In this paper, we introduced the Image-based Event Sequencing (IES) challenge along with the Blocksworld Image Reasoning Dataset (BIRD) that we believe has the potential to open new research avenues in cognition-based learning and reasoning, and as a step towards combining learning and reasoning in computer vision.
Our experiments show that end-to-end deep neural networks fail to reliably generate event-sequences and do not exhibit inductive generalization. 
We argue that encoding scenes from pixel-domain into interpretable representations and then using these for reasoning has an edge over learning to reason directly from pixels.
By decomposing the task into two modules - perception and sequencing, we propose a two-stage approach that has multiple advantages.
Firstly, the sequencing benefits from a perception module that encodes images into meaningful spatial representations.
Next, we show that the sequencing module trained on BIRD can be reused in the natural image domain, by simply replacing the perception module with object detectors.
Finally, our experiments show that modular methods possess inductive generalizability, opening up promising avenues for visual reasoning.
Our future work would deal with expanding BIRD into a more generic dataset, by relaxing constraints on BIRD and making it more generic. We plan to allow a larger variety of actions, a larger set of block characteristics, and also to extend this approach to other complex real-world environments.

\section*{Acknowledgement}
The authors are grateful to the National Science Foundation for Grant 1816039 under the NSF Robust Intelligence Program.

\small
\bibliographystyle{plain}
\bibliography{neurips_2019}

\end{document}